\documentclass[sigconf]{acmart}
\usepackage{subfigure}
\AtBeginDocument{%
  \providecommand\BibTeX{{%
    \normalfont B\kern-0.5em{\scshape i\kern-0.25em b}\kern-0.8em\TeX}}}

\copyrightyear{2020}
\acmYear{2020}
\setcopyright{acmcopyright}\acmConference[SIGIR '20]{Proceedings of the 43rd
International ACM SIGIR Conference on Research and Development in Information
Retrieval}{July 25--30, 2020}{Virtual Event, China}
\acmBooktitle{Proceedings of the 43rd International ACM SIGIR Conference on
Research and Development in Information Retrieval (SIGIR '20), July 25--30, 2020,
Virtual Event, China}
\acmPrice{15.00}
\acmDOI{10.1145/3300000.3400000}
\acmISBN{978-1-4503-XXXX-X/XX/XX}

\begin{document}
\fancyhead{}
\title{Robust Layout-aware IE for Visually Rich Documents with Pre-trained Language Models}

\author{Mengxi Wei}
\email{cangqing.wmx@alibaba-inc.com}
\affiliation{%
  \institution{Alibaba Group}
}

\author{Yifan He}
\email{y.he@alibaba-inc.com}
\affiliation{%
  \institution{Alibaba Group}
}

\author{Qiong Zhang}
\email{qz.zhang@alibaba-inc.com}
\affiliation{%
  \institution{Alibaba Group}
}

\begin{abstract}
\begin{sloppypar}
Many business documents processed in modern NLP and IR pipelines are visually rich: in addition to text, their semantics can also be captured by visual traits such as layout, format, and fonts. We study the problem of information extraction from visually rich documents (VRDs) and present a model that combines the power of large pre-trained language models and graph neural networks to efficiently encode both textual and visual information in business documents. We further introduce new fine-tuning objectives to improve in-domain unsupervised fine-tuning to better utilize large amount of unlabeled in-domain data.

We experiment on real world invoice and resume data sets and show that the proposed method outperforms strong text-based RoBERTa baselines by 6.3\% absolute F1 on invoices and 4.7\% absolute F1 on resumes. When evaluated in a few-shot setting, our method requires up to 30x less annotation data than the baseline to achieve the same level of performance at $\sim90\%$ F1.
\end{sloppypar}
\end{abstract}

\keywords{Visually Rich Document, Structured Information Extraction, \\ Graph Neural Networks}

\maketitle

\section{Introduction}

Information extraction (IE) is the process of identifying within text instances of specified classes of entities as well as relations and events involving these entities~\cite{grishman2012ie}. It is crucial to down stream search and knowledge base applications and has witnessed rapid progress in recent years, especially with the development of neural networks and large pre-trained language models. But IE performance still leaves much to be desired in many real world scenarios. One reason is that most IE research and applications to date assume the input to be text strings, while real world IE systems often have to process business documents with rich visual layout, such as invoices and resumes.

This mismatch limits the IE systems' access to informative visual clues that can help extraction. Consider the examples in Figure~\ref{fig:examples}: the layouts for invoices, resumes, and job ads contain important information: section titles in resumes and job ads are often in fonts different from the content and prices in invoices are often listed in the same column with ``Amount'' as the column head. Such information is ignored by models that rely solely on text information and IE performance is hindered as as result.

Incorporating visual features is therefore crucial for IE applications to effectively process visually rich business documents: first, combining evidence from both the text and the layout can allow the model to make more accurate predictions; second, as more discriminative layout information is encoded into the model as features, such models may require less labeled data to train.


A number of pioneering work has noticed the importance of visual features in information extraction: \cite{DBLP:conf/naacl/LiuGZZ19} combines the Bi-LSTM\cite{DBLP:journals/tsp/SchusterP97} with graph convolution for information extraction from VRDs for scanned images and has shown substantial improvement on performance. \cite{DBLP:conf/naacl/QianSJGB19} introduces GraphIE, which integrates Graph Convolutional Network (GCN~\cite{GCN}) with Bi-LSTM for sequence tagging and uses the GCN to encode visual features. However, they focus on the basic layout information (such as coordinates of text boxes), but ignore richer layout semantics that we explore in this work. In addition, these models do not have access to pre-trained language models.

Recently, LayoutLM \cite{DBLP:journals/corr/abs-1912-13318} adds layout-informed embeddings in addition to text embeddings in pre-training to jointly model text and layout information to better utilize visual information in OCR documents. LayoutLM has shown significant improvement on receipt understanding and form understanding problems in experiments, but it is mainly designed for OCR documents, where the main features are bounding box coordinates. Still, a large portion of today's business documents are digital born and offer richer and more accurate layout information than OCR. We aim to better utilize such information in this work. In addition, even the same type of business documents can have very different visual traits when the domain changes: e.g. resumes for marketing personnels and software engineers can have very different layouts, and invoices from different countries have different patterns. We try to address these problems through more detailed modeling of layout semantics and in-domain fine-tuning.

In this paper, we present a robust entity extraction model for visually rich documents, based on graph networks and pre-trained language models. Different from previous work, we focus on digital-born documents that are more prevalent in industry. We try to combine graph-based IE with pre-trained LMs to make more precise predictions, to facilitate more effective layout-aware in-domain fine-tuning, and to reduce the amount of annotation in zero- and few-shot settings:

\begin{figure*}[h!t]
    \centering
    \subfigure[Invoice]{
    \begin{minipage}[b]{0.25\textwidth}
        \includegraphics[width=\textwidth]{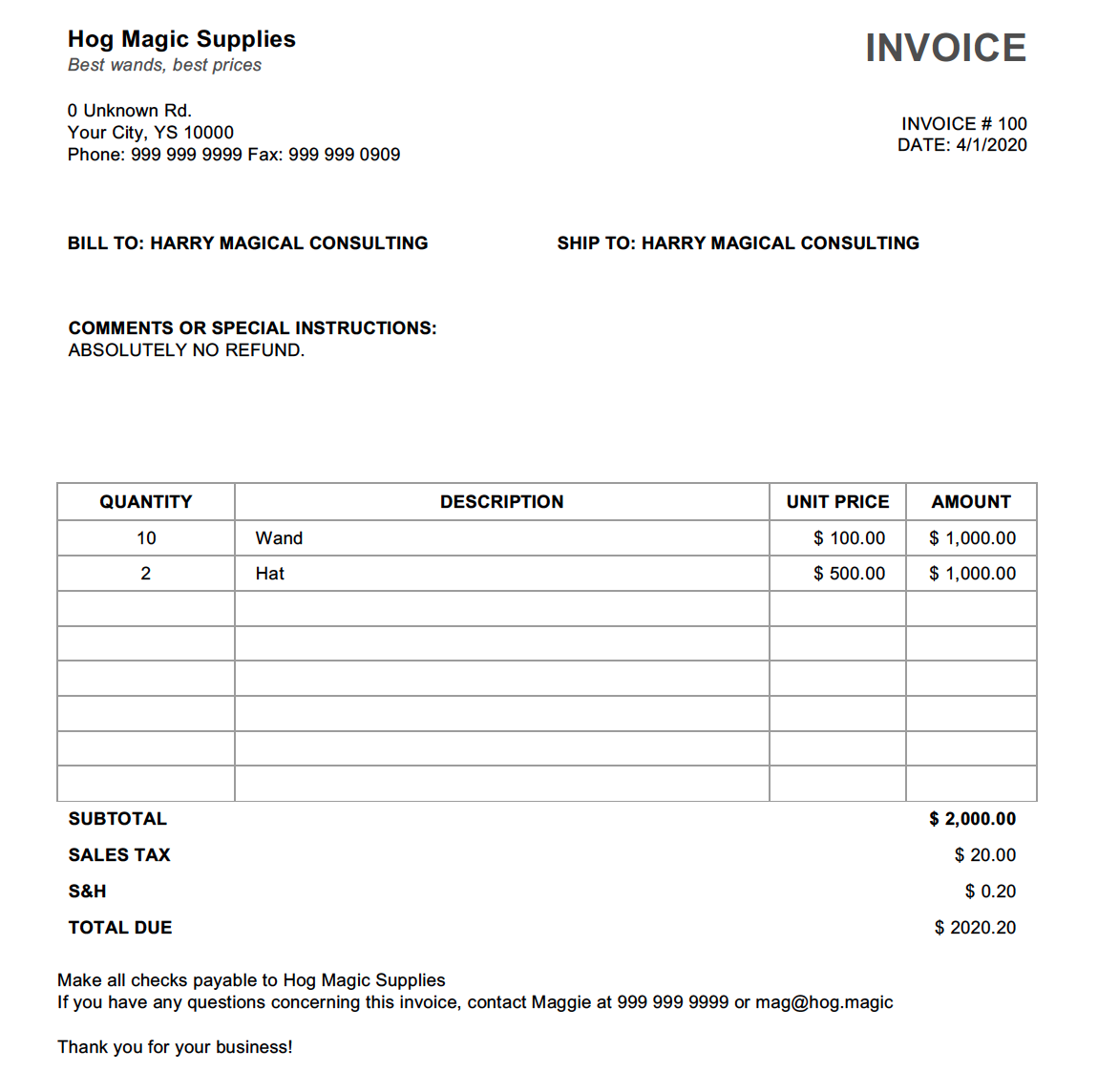}
        \label{invoice}
    \end{minipage}}
    \subfigure[Resume]{
    \begin{minipage}[b]{0.20\textwidth}
        \includegraphics[width=\textwidth]{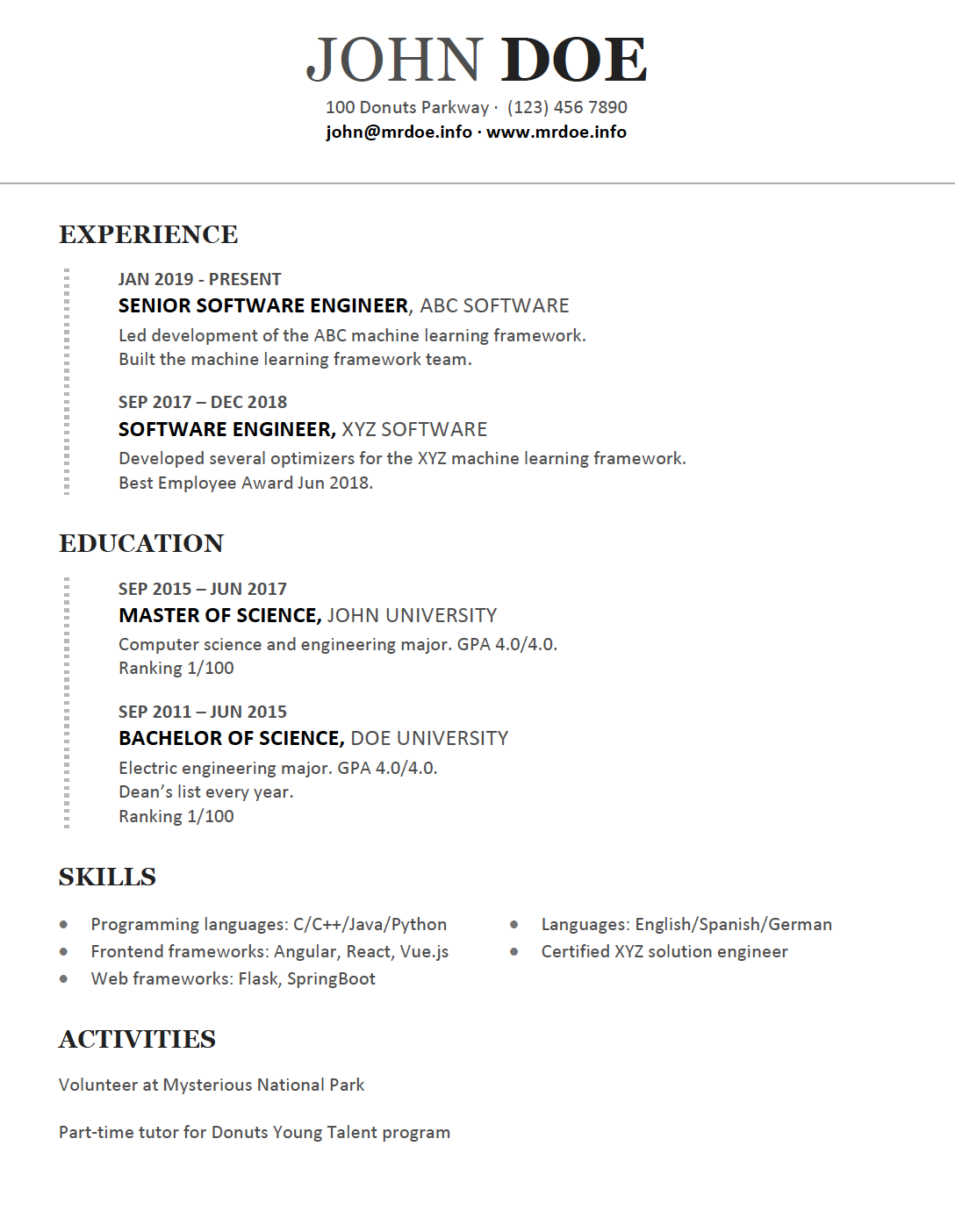}
        \label{resume}
    \end{minipage}}
    \subfigure[Job Ad]{
    \begin{minipage}[b]{0.21\textwidth}
        \includegraphics[width=\textwidth]{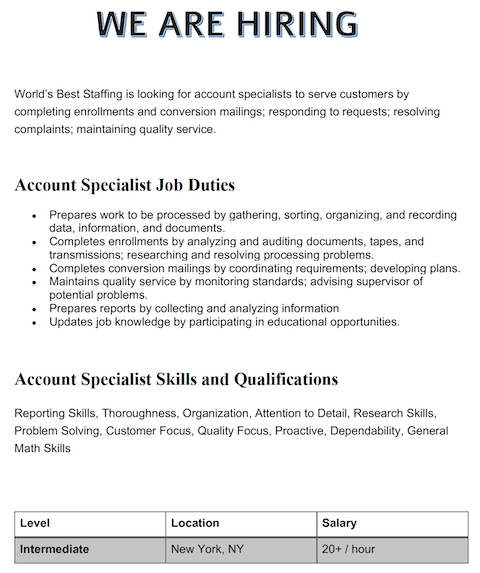}
        \label{notice}
    \end{minipage}}
    \caption{Digital-born documents of different fields in industry. Contents are fictional.}
    \Description{Digital-born documents of different fields in industry. Contents are fictional.}
\label{fig:examples}\end{figure*}

\paragraph{Graph-based IE with pre-trained LMs.} Our baseline text encoder is based on pre-trained transformer-based \cite{DBLP:journals/corr/abs-1810-04805} language models that have recently brought significant performance gains in many NLP tasks. We use RoBERTa~\cite{Liu2019RoBERTa} to extract information from plain text as baseline and use its architecture as our encoder backbone. Then a GCN-based graph module is added on top of the RoBERTa network, so that the visual information of a text box could be encoded into the end-to-end model as edges and nodes. We accommodate different edge types to capture domain specific layout semantics in VSDs and integrate the information of font type and font size into the node to better model non-standard layouts with the graph.

\paragraph{Layout-aware in-domain fine-tuning.}
\begin{sloppypar}
We introduce two fine-tuning objectives: Sequence Positional Relationship Classification (SPRC) and Masked Language Model (MLM) to fine-tune the models on unlabeled in-domain data. SPRC is a four-label sentence classification loss for layout relationship between two adjacent text boxes and aims to force the language model to learn from visual context. MLM is a cross-entropy loss on predicting the masked tokens that aims to force the model to learn the semantic information from context.
\end{sloppypar}
\paragraph{Zero- and few-shot behavior.} We evaluate the model on two real world document information extraction scenarios: invoice analysis and resume information extraction. For invoice analysis, we collect a large number of real-world international invoices from different departments within a multinational corporation. The departments have different vendors with different invoice format. We both test the overall performance of our system on a portion of the dataset and reserve invoices from a small set of departments for zero- and few-shot experiments.

In addition, we apply our models to a collection of English resumes for a wide variety of professions. The resumes are different from invoices as they are more text-centric and have more flexible formats.

On both data sets, our model outperforms strong RoBERTa baselines with significant margins: for the invoice dataset, the model with GCN outperforms the baseline by about 4.8\% F1, and the model fine tuned with two proposed unsupervised fine-tuning tasks achieves the best performance with 6.3\% F1 improvement at 95.87\%. For the resume dataset, the model with GCN and two fine-tuning objectives outperforms the RoBERTa baseline by 4.7\%.

In summary, the contributions of this study are as follows:
\begin{itemize}
\begin{sloppypar}
\item To the best of our knowledge, it is the first model that combines a graph neural network module with pre-trained transformer-based language models to encode both text and rich layout information in VRDs. Although the pre-trained language model already achieves strong performance on text-based tasks, combining it with a graph-based layout encoder still achieves a significant improvement for visually rich documents.
\end{sloppypar}
\item Two training tasks are proposed to fine tune the model without labeled data, which further improves the extraction performance without using additional training data.
\item We conduct extensive experiments on different datasets and report positive results on zero-shot and few-shot scenarios, where annotated instances are scarce.
\end{itemize}

In the rest of the paper, we first review related work ($\S$~\ref{sec:relatedwork}) and then describe our model in detail ($\S$~\ref{sec:methodology}), including the text ($\S$~\ref{sec:methodtext}) and layout encoders ($\S$~\ref{sec:methodgraph}) in our model, as well as the unsupervised fine-tuning tasks($\S$~\ref{sec:methodpretrain}). We report experimental results on invoices and resumes ($\S$~\ref{sec:experiments}), in both supervised and zero- and few-shot settings.

\section{Related Work}\label{sec:relatedwork}
\subsection{IE for Visually Rich Documents}
\begin{sloppypar}
Our work falls under the scope of Visually Rich Document information extraction which is a relatively new research topic. We roughly divide the current progress of the approaches on this problem into three categories. The first category is rule-based systems. \cite{rusinol2013field} describes an invoice extraction system using a number of rules and empirical features, and \cite{d2018field} builds a more stable system using tf-idf algorithm and a large number of human-designed features. Other document-level entity extraction systems combine rules and statistical models \cite{belaid2001adaptive,schuster2013intellix}. Although it is possible to craft high-precision rules in some closed-domain applications, rule-based systems are usually associated with extensive human effort and cannot be rapidly adapted to new domains.

The second category is graph-based statistical models, where graphs are used to model the relationships between layout components, such as text boxes. \cite{santosh2013pattern} first performs graph mining in a document with a set of key-fields
selected by clients, in order to learn the pattern to extract information in the absence of clients. More recently, graph neural networks are used to capture structural information in visually rich documents: \cite{DBLP:conf/naacl/LiuGZZ19} applies graph modules to encode the visual information with deep neural networks and GraphIE~\cite{DBLP:conf/naacl/QianSJGB19} also assumes that the graph structure is ubiquitous in the text, and applies GCN between the BiLSTM encoder-decoder structure to model the layout information in the document. The limitation of these methods is that they do not have access to pre-trained language models such as BERT and have not explored rich visual information (e.g. font and weight of texts) beyond the position of texts.

The third category is approaches that exploit 2D grid information of characters or words. Chargrid~\cite{katti2018chargrid} models the problem by encoding each document page as a two-dimensional grid of characters and used a fully convolutional encoder-decoder network to predict the class of all the characters and group them by the predicted item bounding boxes. LayoutLM~\cite{DBLP:journals/corr/abs-1912-13318} appends the language model embeddings with 2D grid information of words to jointly pre-train text and layout. These methods rely on the results of OCR and only model with information on the character and word levels, while valuable sequence-level information in digital-born documents is ignored.
\end{sloppypar}
\subsection{Graph Neural Networks}
\begin{sloppypar}
Graph neural networks have attracted increasing attention. GCN~\cite{GCN} models graph-structured data based on an efficient variant of convolutional neural networks, which has been proved effective in many NLP tasks. For example, GraphRel\cite{fu2019graphrel} uses a Bi-GCN to better extract relations by jointly learning named entities and relations. \cite{liu2018matching} models and matches long article through a graph convolutional network to identify the relationship between two articles. \cite{marcheggiani2017encoding} encodes sentences with GCN for semantic role labeling. In our model, GCN architecture is applied to encode the positional and formatting relationship between the text nodes.

Attention mechanisms have been added to graph neural networks for more expressiveness. GAT~\cite{GAT} enables specifying different weights to different nodes and outperforms its convolutional counterparts on knowledge graph dataset. Then Graph Transformers\cite{Gtrans} present an end-to-end system for graph-to-text generation from knowledge graphs with multi-head attention inspired by transformers.
\end{sloppypar}
\subsection{Pre-trained Transformer-based Language Models}
Since BERT~\cite{DBLP:journals/corr/abs-1810-04805} achieved SOTA performances on a range of NLP tasks, there has been much more attention on pre-training with large amounts of unlabeled data to produce powerful contextual representations for downstream tasks. After pre-training from massive text data with language modeling objectives~\cite{ULMfit,Elmo,GPT}, the encoder could learn high-level dependencies between tokens and then feed the representations to downstream tasks to perform transfer learning. In addition to language model related objectives, many works apply other pre-training objectives which has also made significant progress. BERT introduces a next sentence prediction (NSP) loss considering the sentence level relationships. \cite{mayhew2019robust} comes up with a pre-training objective predicts casing in the text to address the robustness problem for named entity recognition (NER) systems and \cite{mehri2019pretraining} introduces four different pre-training objectives for dialog context representation learning.

In our work, we fine tune the proposed RoBERTa-GCN model with two objectives, the Masked Language Model (MLM) used by BERT and RoBERTa, as well as the novel sequence positional relationship classification loss (SPRC), which is designed to capture the relationships between text chunks given their position and content.

\subsection{Few Shot Learning}
\begin{sloppypar}
Training statistical models typically requires annotating a large amount of data, which can often be costly. As a remedy, the few shot learning task~\cite{miller2000learning,fei2006one} is proposed to build models that can quickly learn from a small number of samples. To address this problem, our method should be robust enough to adapt to an unseen dataset based on very few samples. Many few-shot learning methods consider image domains e.g.~\cite{vinyals2016matching}. For NLP, \cite{DBLP:conf/naacl/YuGYCPCTWZ18} proposes an adaptive metric learning approach to automatically assign weights for newly seen few-shot task and \cite{geng2019few} uses a few-shot learning method that leverages the dynamic routing algorithm in meta-learning for intention recognition in conversation.

Our approach of binding the GCN with pre-trained BERT-like model and fine-tuning the model with in-domain unlabeled data facilitates few-shot learning by optimizing both language features (through pre-trained transformer-based language models) and visual features (through layout-aware fine-tuning on unlabeled in-domain data) on  unlabeled data as much as possible. Experiments show that our method outperforms strong text-based baseline by significant margins in zero- and few-shot settings.
\end{sloppypar}
\section{Methodology}\label{sec:methodology}
\begin{figure}[h]
  \centering
  \includegraphics[width=0.7\linewidth]{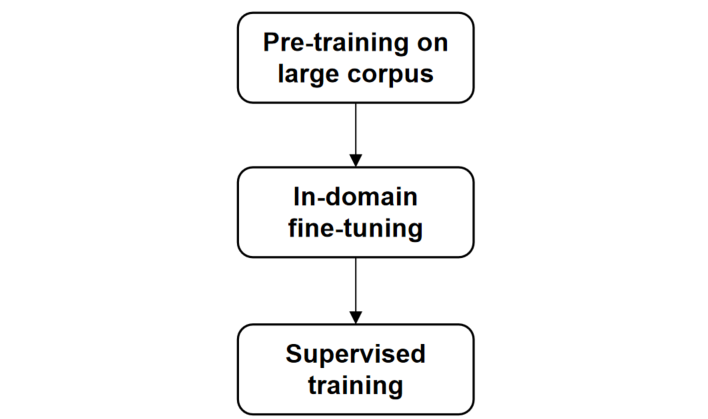}
  \caption{Transfer learning steps for pre-trained network}
  \Description{The transfer learning steps for the language model.}
\label{fig:tansferlearning}\end{figure}

\begin{figure*}[h]
  \centering
  \includegraphics[width=0.9\linewidth]{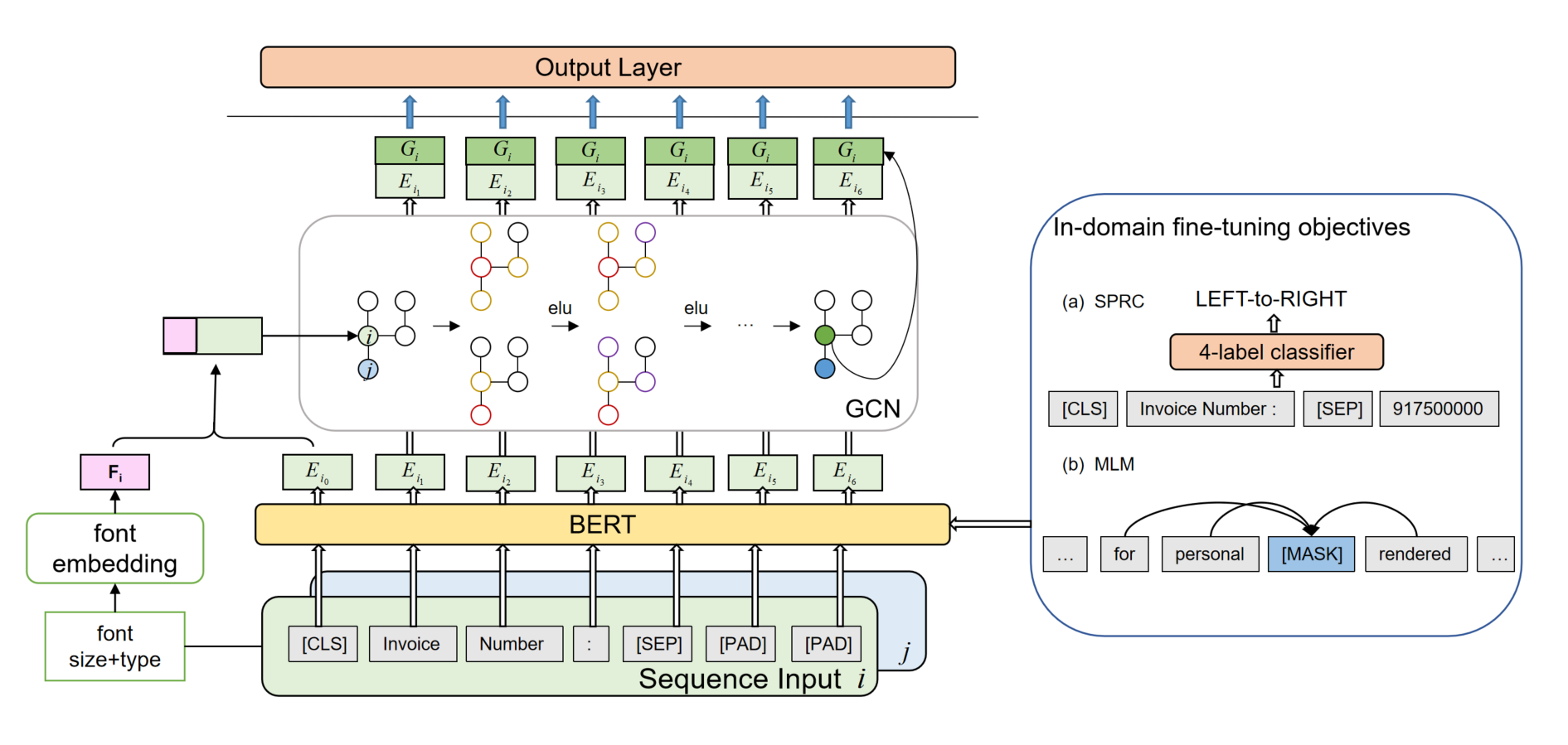}
  \caption{Model architecture}
  \Description{The main architecture.}
\label{fig:main}\end{figure*}

In this section, we present our document information extraction model architecture and pre-trained methods. The basic training procedure is to use a powerful transformer-based language model pre-trained on a large corpus as the encoder backbone, followed by fine-tuning the encoder with unsupervised objectives on in-domain knowledge. The final step is to train the entity extraction model with labeled data. The procedure is illustrated in Figure~\ref{fig:tansferlearning} and the overall architecture of the proposed model is illustrated in Figure~\ref{fig:main}.

As shown in Figure~\ref{fig:main}, we use transformer-based language models to encode the text, and the GCN to encode layout and positional information. We initialize the weight of LM encoder with the pre-trained RoBERTa-BASE and fine tune the encoder on unlabeled in-domain data with two objectives: Masked LM as in RoBERTa and a novel Sequence Positional Relationship Classification (SPRC) objective.
\subsection{Problem Setup}
We focus on digital-born documents, which are documents produced and received in digital form without printing and OCR. These documents preserve rich formatting information and constitute a large portion of business documents to be analyzed today. We model each document as a set of text boxes and apply sequence tagging on each text box. Each text box corresponds to a node in the graph. We create a graph for each page, if a document has multiple pages.

To convert a document page to a graph, we define the graph as G, where $ G=(V,E) $, $V =\{v_1,v_2,\cdots,v_N\}$ is the set of text boxes in one page, and matrix $E \subset M \times V \times V$ is the undirected edges set, where M is the number of edge types. In our experiments, we define a basic undirected edge type $e_{i,j} = (v_i,v_j)$ that connects a text box to its closest vertical or horizontal neighbor. The model can also accommodate multiple edge types defined in different domains (cf. \S \ref{sec:methodgraph}).

Our model combines text features and layout features to perform entity extraction.

\subsection{Pre-trained LM Text Encoder}\label{sec:methodtext}
Different from some existing works~\cite{DBLP:conf/naacl/LiuGZZ19,DBLP:conf/naacl/QianSJGB19} that encode words or sentences using the standard BiLSTM-CRF\cite{DBLP:journals/tsp/SchusterP97}, we take advantage of pre-trained transformer-based language models influenced by BERT~\cite{DBLP:journals/corr/abs-1810-04805,Liu2019RoBERTa} to perform text feature extraction as they are much more expressive.

The BERT model is based on the multi-layer bidirectional transformer \cite{DBLP:conf/nips/VaswaniSPUJGKP17} architecture that effectively encodes a sequence of tokens to produce final representations of the text. It is often trained on large unlabeled text data using a language model objective and then fine-tuned on in-domain data to effectively transfer knowledge from the large-scale unlabeled text corpus for a specific task.

The BERT model takes as input a set of tokens. For each token, its vector representation is computed by summing the corresponding token embedding, positional embedding, and segment embedding. A special classification embedding [CLS] is always added as the first token of the sequence, and a special end-of-sequence token [SEP] is added to the end. Then, these inputs are passed through a multi-layer bidirectional transformer encoder.

During pre-training, BERT uses two objectives: Masked Language Model (MLM) and Next Sentence Prediction (NSP). The MLM objective randomly masks tokens in a sequence and the language model is trained to recover the masked tokens. The NSP objective performs binary classification on whether two segments follow each other in the original text. As it is pre-trained on a large corpus with a large number of parameters, BERT-like language model has shown excellent transferability~\cite{peng2019transfer,DocBERT,nogueira2019passage}: downstream tasks can often achieve high accuracy even if the data size of the specific task is not large enough, when the models are initialized with pre-trained BERT weights.
\begin{sloppypar}
In our work, we use the pre-trained RoBERTa model~\cite{Liu2019RoBERTa}, which is similar to BERT in architecture but tuned with different objectives on larger data, as the encoder backbone to obtain expressive representations of text. Given a tokenized text box $s_i = (\omega_1^{(i)},\omega_2^{(i)},\cdots,\omega_k^{(i)})$ of length $k$ as input, we pool the RoBERTa model output by concatenating the final hidden state corresponding to the first token [CLS] as the aggregate representation of the sentence. The sentence representation vector [CLS] is used as messages in the graph neural network to model page layout on the graph level. The other token embeddings $H_{1:k}^{(i)}$ will be concatenated with the outputs of the graph layers (cf. \S \ref{sec:methodgraph}) as features for the entity tagger.
\end{sloppypar}
\begin{equation}
    H_{0:k}^{(i)} = RoBERTa\left(\omega_{0:k}^{(i)};\Theta\right)
\end{equation}
\begin{equation}
    C_i = H_0^{(i)}
\end{equation}

\subsection{Graph-based Layout Encoder}\label{sec:methodgraph}
Empirically, visual information of non-standard format VRDs varies greatly in different scenarios. For example, information in invoices is often shown as lists, while information in resume is often organized in sections. Accordingly, we adapt the graph structure to model the sophisticated layouts and fonts to learn task specific visual features, rather than relying solely on the bounding box positions into the pre-trained model like \cite{DBLP:journals/corr/abs-1912-13318}.

\subsubsection{Node-level Features}

To convert sentences in one document page to a graph, we define each graph node $v_i$ as the RoBERTa output $C_i$ of a text box along with its fonts encodings.

We define font encodings as follows. We use the font name and font size to represent a specific type of the text font, and sort all font types appearing in one document from high frequency to low frequency. Then the font is numbered by its rank. Formally, assume for text box $s_i$, the font type of the text is defined as $f_i$, the input of node $v_i$ for the first graph layer is defined as $h_i^0$,
\begin{equation}
    h_i^0 = C_i||e\left(f_i\right)
\end{equation}
\noindent
where $||$ is the concatenate operation, $e(\cdot)$ is an embedding lookup function. Embeddings are initialized randomly. The intuition is the separate the font used for content from the font used for headers.

\subsubsection{Graph Convolution Network}
We use Graph Convolutional Network (GCN)~\cite{GCN} to model the graph context in our model, where nodes encode text-box level text and format information and edges capture the layout information of the whole page. A GCN layer convolves the features of a node's nearest neighbors, and multiple layers can model more complex relationships between nodes. Node representations of the last GCN layer are used as layout features for entity extraction.

Each node aggregates the information from neighboring node's features with a GCN layer as follows,
\begin{equation}
    h_i^{l+1} = eLU\left(\frac{1}{N}\sum_{j\in N(i)}\left(W^lh_i^l+b^l\right)\right)
\end{equation}
\noindent
where $N(i)$ represents the nodes connect to $i$, including $i$ itself. $N$ is the size of $N(i)$, $h_i^l$ is the representation of node $i$ at the $l$-th layer. $W^l$ is the weight, $b^l$ is a bias parameter, and the activation function $eLU$ is the exponential linear unit\cite{elu} which obtains better performance than other activation functions in initial experiments.

Text boxes on a page may have different types of relationships. Some relationships are spatial (e.g. the closest text box on the left), while others are based on formatting (e.g. the closest text box using a larger font). Our model can handle different types of relations between nodes. In particular, when there are multiple types of edges, Eq.(4) is modified as:
\begin{equation}
    h_i^{l+1} = eLU\left(\frac{1}{N_t}
    \frac{1}{N}\sum_{t=1}^{N_t}\sum_{j\in N(i)}\left(W_t^lh_i^l+b^l\right)\right)
\end{equation}
where $W_t$ is the weight for $t$-th type of edges and $N_t$ is the number of edge types.

Skip connection\cite{DBLP:conf/cvpr/HeZRS16} is also applied for multiple GCN layers to keep the information from previous layers and reduce overfitting:
\begin{equation}
     h_i^{l+1} = eLU\left(\frac{1}{N_t}
        \frac{1}{N}\sum_{t=1}^{N_t}\sum_{j\in N(i)}\left(W_t^lh_i^l+b^l\right)+h_i^l\right)
\end{equation}

We also experiment with GAT\cite{GAT} and G-Trans\cite{Gtrans} layers in initial experiments, but results show worse performance than GCN.

\subsection{Entity extraction}
With the layout encoding from GCN and the text encoding from the pre-trained language model, we sequentially tag each word in a sentence to extract entities. As illustrated in Figure\ref{fig:main}, the final layer output representation of the i-th text box $h_i^L$ from the GCN is concatenated to each token states $H_{1:k}^{(i)}$ from the RoBERTa encoder. Then, a sequence labeling layer is used to predict entity types for each token. Entity types are encoded at token level with the BIO schema.

\subsection{Fine-tuning objectives}\label{sec:methodpretrain}
Labeling data is expensive for information extraction applications, but unlabeled data is more likely abundant. So we try to make full use of the large amount of unlabeled data in our model through unsupervised in-domain fine-tuning.

Specifically, assuming that we have data $D$ for a given task. In addition to training all labeled document data $D_{labeled}$, we also make use of plenty of unlabeled data $D_{unlabeled}$ during the fine-tuning process to adapt the pre-trained model to a new task.

An extraction task can cover different domains. Invoice extraction, for example, may have to process both hotel invoice and appliance invoice for different users. Document from different domains often use different language and layout and it is traditionally expensive to adapt models trained on one domain to other domains. To better evaluate the performance of our system on different domains, we further divide $D_{labeled}$ into $D_{seen}$ and $D_{unseen}$, where $D_{unseen}$ is a set of testing documents from unseen domains. The model is first fine-tuned with $D_{unlabeled}$, and then trained to extract entities based on a portion of $D_{seen}$. It is finally tested on the test portion of $D_{seen}$ and $D_{unseen}$.

Inspired by existing pre-trained models, we propose to fine tune on unlabeled in-domain data with two training objectives before supervised training. Experiments show that in-domain fine-tuning helps extraction performance on both $D_{seen}$ and $D_{unseen}$.

\subsubsection{Sequences Positional relationship classification}
In addition to the graph module which is designed to represent the information of document layout, we attempt to utilize layout information during the fine-tuning process as well. We propose a fine-tuning objective named sequences positional relationship classification (SPRC) to enrich the language model representation with some basic layout information.

Specifically, we extract strictly adjacent text pairs from the unlabeled dataset $D_{unlabeled}$ to predict which of the following relationship types they belong to: left-right, right-left, up-down and down-up. Strict adjacency means that the two text boxes should share either the same x- or y-coordinate, so that at least one of edges of the two boxes is aligned. Two sentences are concatenated as input with a [SEP] token between them and the sentence encoding [CLS] is fed into a linear layer to predict the relation type.

This task is designed to capture domain-specific layout information from unlabeled VRDs: for example, if a proper noun is aligned to a position name such as "Software Engineer", it is likely that the proper noun is a company name. When fine-tuning with the SPRC objective, the language model becomes informed of which terms are correlated according to the layout.

\subsubsection{Masked Language Model}
We keep the MLM training objective of BERT model to fine-tune on $D_{unlabeled}$ to get better representation of each token for our task. During the training, some of the input tokens are randomly masked and the model is trained to predict these masked tokens given the contexts. In the original BERT implementation, masking is performed during data preprocessing which resulting in a static mask. We fine-tune our task with \cite{Liu2019RoBERTa}'s MLM implementation that use a dynamic masking strategy, which generates the masking patterns during training.

In addition to using these two fine-tuning objectives separately, we also experiment with fine-tuning sequentially on the MLM and the SPRC objectives, and then use the fine-tuned model to perform supervised training on $D_{labeled}$.
\section{Experiments}\label{sec:experiments}
We evaluate our model and fine-tuning methods for layout-aware information extraction with a real-world VRD task: invoice entity extraction. The invoices are collected from a multi-national corporation with a number of departments. Each department has its own set of vendors. We reserve data from some of the departments for testing in a few-shot learning setting (as $D_{unseen}$, cf. $\S$~\ref{sec:methodpretrain}). To demonstrate the generality of our proposed approach, we finally experiment with another real-world dataset in a completely different field: resume information extraction.

\subsection{Invoice Information Extraction}
\subsubsection{Dataset and Pre-processing}
\begin{figure}[h]
  \centering
  \includegraphics[width=\linewidth]{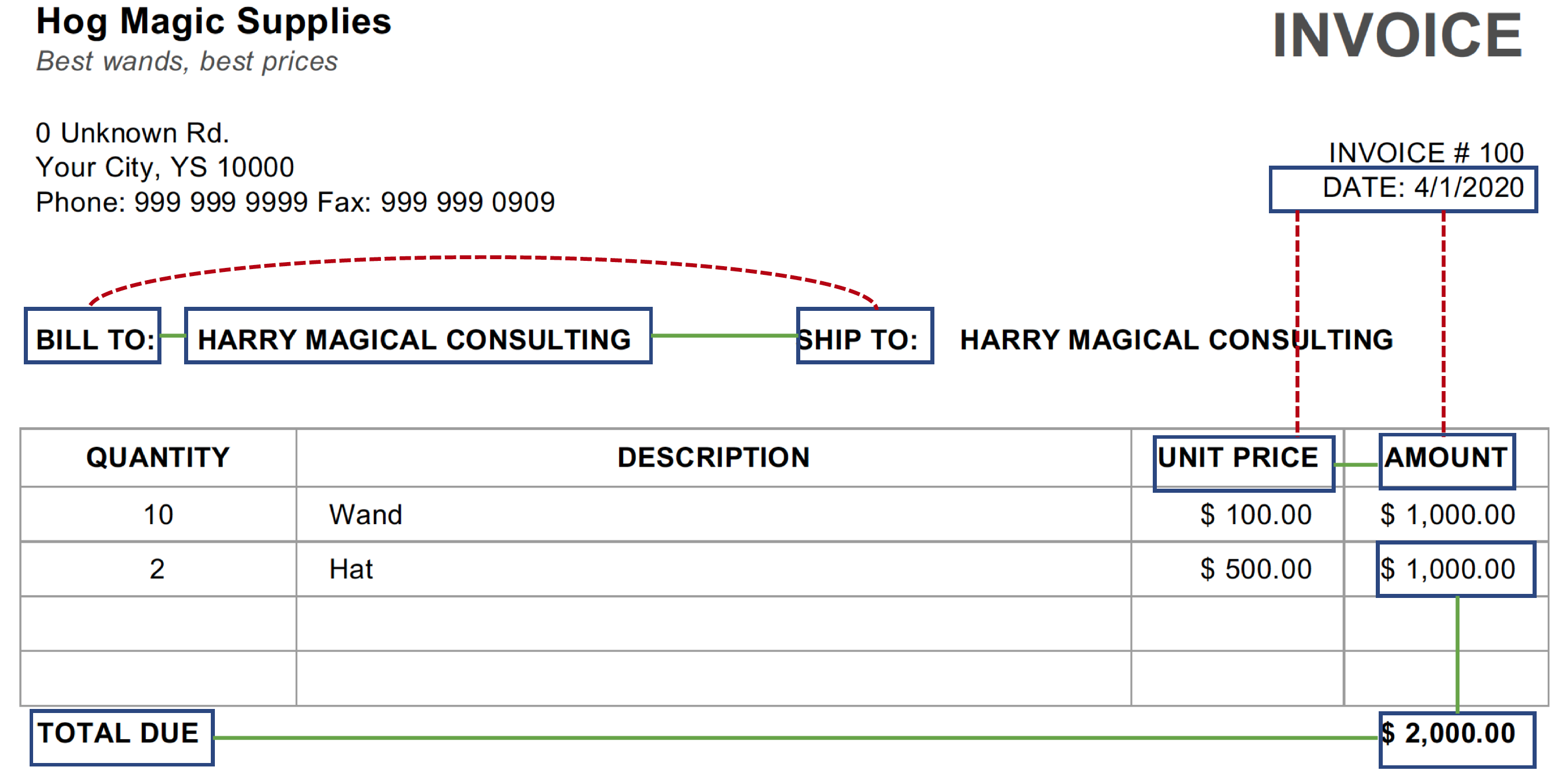}
  \caption{An invoice example.}
  \Description{An invoice example.}
\label{fig:invoicesample}\end{figure}
The invoice dataset we use consists of 10,450 real-world international invoices which are in digital-born pdf format. We collect the dataset from over sixty departments of an international company. Invoices from the same department may be provided by the same vendor and therefore share the same template and layout, but generally, these invoices are in different non-standard formats and exhibit large variability in the layout.
\begin{table}
  \caption{Invoice Dataset Statistics}
  \label{tab:invoicedata}
  \begin{tabular}{lcc}
    \toprule
    Dataset&page&text box\\
    \midrule
    $D_{labeled}$ & 2,000 & 69,560\\
    $D_{seen}$ & 1,857& 64,130 \\
    $D_{unseen}$ & 143 & 5,430\\
    \midrule
    $D_{unlabeled}$ & 10,405& 369,927\\
  \bottomrule
\end{tabular}
\end{table}
As shown in Table~\ref{tab:invoicedata}, we have 2,000 manually labeled invoices, $D_{labeled}$ ($D_{labeled}\subset D_{unlabeled}$), which have four types of entities: SellerName, PurchaserName, InvoiceNo and Amount. For experiments in zero- and few-shot settings, we reserve the invoices in a specific department, whose sellers are different from other departments, as an unseen dataset $D_{unseen}$, where $D_{labeled} = D_{unseen}\cup D_{seen}$.

For pre-processing, we use a text extraction tool for PDF documents,  PDFMiner~\footnote{https://euske.github.io/pdfminer}, to first covert PDFs into text boxes along with their bounding-box positions and font information. We also use rules to merge boxes that are extremely close to each other and assume that each text box represents a complete word or sentence: i.e. if the text box contains an entity, it contains at least one complete entity. Then a graph module is used to encode a document page and each text box corresponds to a graph node.

\subsubsection{Baselines}\label{sec:experimentbaselines}
We compare our system with several strong baselines that only take text input, including BiLSTM-CRF and pre-trained transformer-based models, which are all widely used for sequence tagging.

The standard BiLSTM-CRF network we use takes each text segment as an input, and tokens are encoded with the word embeddings vectors. We also use a character-level Bi-LSTM layer which takes embedded characters as inputs to learn character-level features. The output hidden states of the character Bi-LSTM are concatenated with word embeddings and then fed into a Bi-LSTM layer and a CRF layer. The word embeddings we use are initialized with pre-trained GloVe vectors~\footnote{http://nlp.stanford.edu/projects/glove/} while the character embeddings are randomly initialized.

We also compare our model with two SOTA pre-trained models: BERT and RoBERTa. These two models share the same text input and encoder backbone with our method and we reimplement them with the PyTorch Transformers package~\cite{Wolf2019HuggingFacesTS}. RoBERTa shares the same architecture with BERT but is different in pre-training objectives, which achieves better results by removing the NSP objective and dynamically changing the masking patterns. In our experiments, the BASE architecture with a 768 hidden sizes 12-layer Transformer and 12 attention heads is used for both BERT and RoBERTa.

\begin{sloppypar}
From Table~\ref{tab:invoicedataf1}, we observe that transformer-based pre-trained models perform better than Bi-LSTM model, and RoBERTa obtains slightly better F1-score than BERT. Based this observation, we choose RoBERTa to be the encoder backbone of our model.
\end{sloppypar}

\subsubsection{Experimental Setup}
For the edges in a graph, we connect one text box with its closest horizontal or vertical neighbor (cf. $\S$~\ref{sec:methodgraph}). We only connect the adjacent pairs that have the same x or y coordinate. Take text boxes in Figure~\ref{fig:invoicesample} as an example, a green solid line connecting two boxes indicates that there is a graph edge between them, while the red dotted lines indicate that there are no edges. As for other settings of the graph, we use 2-layer GCN with 256 hidden dims and skip connection and an 8-dimensional font embedding. To limit the size of the model, we set the maximum node number to 100 and the maximum sentence length to 50.

For the encoder backbone RoBERTa, we use the RoBERTa-BASE model (cf. $\S$~\ref{sec:experimentbaselines}). The dropout ratio for all layers is set to 0.1, and the learning rate for BERT is set to 1e-5 while for the GCN and the linear classification layer is set to 5e-5. The model is trained with Adam~\cite{kingma2014adam}. We skip the warm-up step because it does not show improvements in initial experiments.

For the two fine-tuning tasks, we set learning rate to 5e-5 for classifier and 1e-5 for RoBERTa.

\subsubsection{Evaluation}
We evaluate our model and the baseline methods using~\textbf{F1-score}. The F1 score is known as a measure of a test's accuracy which considers both the precision and the recall. Precision is the percentage of entity tokens identified by the system that are correct, while recall is the percentage of entity tokens present in the gold annotation that are identified by the system.
\begin{equation}
    Precision=\frac{TP}{TP+FP}
\end{equation}
\begin{equation}
    Recall=\frac{TP}{TP+FN}
\end{equation}
\begin{equation}
    F_{1}=\frac{2\times Precision\times Recall}{Precision + Recall}
\end{equation}
where $TP$,$FP$,$FN$ stand for true positives, false positives, false negatives respectively.
\subsubsection{Results}
We apply the graph module to model the complicated visual layout of a document page and two fine-tuning objectives to improve the performance of the language model. For comparison, we present the results of our proposed methods in Table~\ref{tab:invoicedataf1} along with the baselines.

\begin{sloppypar}
The results indicate that models utilizing the layout information with graph neural network modules outperform every baseline by significant margins: adding the GCN module without fine-tuning improves F1 from 89.58 to 94.37. Furthermore, fine-tuning the model with SPRC and MLM objectives each improves F1 by about 1 point absolute, and fine-tuning with MLM and then SPRC objective outperforms others with the highest F1-score of 95.87.
\end{sloppypar}

\begin{table}
  \caption{Model accuracy on Invoice Dataset}
  \label{tab:invoicedataf1}
  \begin{tabular}{llc}
    \toprule
    Model&Fine-tuning task&F1\\
    \midrule
    BiLSTM-CRF & - & 88.18\\
    BERT & - & 89.07\\
    RoBERTa & - & 89.58\\
    \midrule
    RoBERTa+GCN & - & 94.37\\
    \midrule
    RoBERTa+GCN & +SPRC & 95.25\\
    RoBERTa+GCN & +MLM & 95.66\\
    RoBERTa+GCN & +MLM+SPRC & 95.87\\
  \bottomrule
\end{tabular}
\end{table}

We present the statistics and performance for all types of entities in labeled dataset in Table~\ref{tab:invoicetagdata} for more detail. Usually, as is shown in Fig~\ref{fig:invoicesample}, there are multiple prices in the invoice document that would confuse the sequence model to distinguish which price means the total amount. Through the results, it is evident that the model with GCN outperforms baseline in all the entities, and the improvement is more significant for \textit{Amount} (a 21.87\% improvememt) which strongly suggests that the graph module could effectively use layout information to help sequence models to extract entities defined with visual information.

\begin{table}
 \newcommand{\tabincell}[2]{\begin{tabular}{@{}#1@{}}#2\end{tabular}}
  \caption{Statistics of the entities in labeled Invoice Dataset}
  \label{tab:invoicetagdata}
  \begin{tabular}{lcccc}
    \toprule
     & \tabincell{c}{Seller\\Name}  & \tabincell{c}{Purchaser\\Name} & InvoiceNo & Amount\\
    \midrule
    entities num  & 1,557 & 2,376 & 1,999& 1,873\\
    RoBERTa F1  & 90.69 & 95.47 & 90.18 & 63.54\\
    RoBERTa+GCN F1 &  94.01 & 97.55 & 94.49 & 85.41\\
    \bottomrule
\end{tabular}
\end{table}

For details of fine-tuning, we first initialize the weights of the encoder backbone of our model with the RoBERTa-BASE model, then apply SPRC and MLM as training objectives separately to fine tune on in-domain data. For the SPRC task, a four-label sentence classification loss is used to predict the relationship type of strictly adjacent text pair selected from $D_{unlabeled}$. We fine tune the model with 15 epochs and obtain accuracy at 87.5 on validation set. A cross-entropy loss on predicting the masked tokens is used for MLM task. We concatenate all the origin text in $D_{unseen}$ feeding into the network and train with 30 epochs where the perplexity (Perplexity score is a measurement of how well a probability model predicts a test data and the lower the better~\cite{JurafskyMartin}.) is 1.9. We also use the fine-tuned weights of MLM task to initialize the SPRC task. The accuracy is 89.8 after 18 epochs. The results of fine-tuning objectives are reported in Table~\ref{tab:invoiceprescores}.

\begin{table}
  \caption{Fine-tuning task scores of Invoice Dataset}
  \label{tab:invoiceprescores}
  \begin{tabular}{lcc}
    \toprule
    Fine-tuning Task& Metric & Score\\
    \midrule
    MLM & Perplexity & 1.9 \\
    SPRC & Accuracy & 87.5 \\
    MLM-SPRC & Accuracy & 89.8 \\
  \bottomrule
\end{tabular}
\end{table}
\subsection{Few Shot Experiment}
Although we have collected a large number of invoices from different sources to build the dataset, the number of document layout types is still limited and it is impractical to prepare labeled invoice document from all domains, so it is necessary to evaluate our method in zero- and few-shot settings to verify the robustness of our model on new domains. Zero-shot learning is well-known as a problem where no in-domain labeled training data is available but the system has to make predictions. Few-shot learning is to test the ability of the neural networks on unseen dataset when provided only a very small number of training instances.

In zero-shot experiments, we select invoices from a specific department from labeled dataset as $D_{unseen}$ as test set and use $D_{seen}$ for training and validation. For few-shot learning, we select 70 invoices from the same department with $D_{unseen}$ from $D_{unlabeled}$ and manually labeled this set as $D_{few}$ to fine-tune our model. It is worth mentioning that $D_{few}\cap D_{unseen}=\emptyset$, $D_{few}\cap D_{labeled}=\emptyset$, which means that the $D_{few}$ is constructed to be used for few-shot experiments only.

Table~\ref{tab:unseeninvoicedataf1} illustrates the results of these experiments. In zero-shot experiments, the performances of both methods are not as good as the supervised setting, where abundant training data is available. Nevertheless, our model still obtains a 3.9\% improvement over the baseline on $D_{unseen}$.

We then focus on the few-shot setting where there are few labeled data fine-tuning the trained model, as a few shot experiment. In our experiment, $D_{few}$ is further splitted into 50 invoices as validation and 20 for fine-tuning. After 5 epochs fine-tuning process, out model outperforms the baseline by a significant margin at 94.62, which is comparable to the performance of fully supervised training. The few-shot results further prove the effectiveness and robustness of our method.

\begin{figure}[h]
  \centering
  \includegraphics[width=0.8\linewidth]{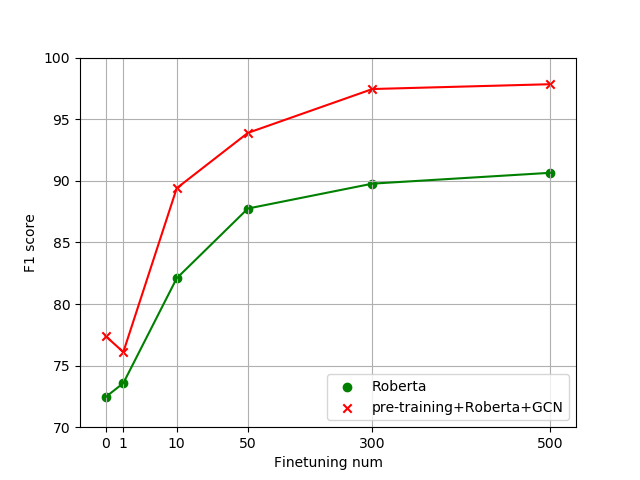}
  \caption{Fine-tuning with different number of unseen data}
  \Description{Fine-tuning with different number of unseen data.}
\label{fig:fewshot}\end{figure}

\begin{sloppypar}
To understand how the system performs as the number of training instances increase, we manually label 500 invoices from the $D_{unseen}$ subset and select different number of training instances (1, 10, 50, 300, and 500) to fine-tune the model. The results on $D_{unseen}$ are illustrated in Figure~\ref{fig:fewshot}. It can be seen from the figure that our model achieves $\sim 90\%$ F1-score with only about ten labeled instances. In contrast, to achieve the same level of performance as our model, the baseline model requires about 300 instances, which translates into 30x annotation cost reduction for the proposed model. We conjure that the improvement is facilitated by our models' ability to effectively utilize both text and layout information in both supervised and unsupervised settings.
\end{sloppypar}

\begin{table}
  \caption{Model accuracy on unseen Invoice Dataset}
  \label{tab:unseeninvoicedataf1}
  \begin{tabular}{cllc}
    \toprule
    Modality&Model&Fine-tuning task&$D_{unseen}$ F1\\
    \midrule
    zero shot &RoBERTa & - & 75.61\\
    zero shot &RoBERTa+GCN & - & 79.59\\
    \midrule
    few shot &RoBERTa & - & 86.42\\
    few shot &RoBERTa+GCN & - & 94.62\\
    \midrule
    few shot &RoBERTa+GCN & +MLM+SPRC & 95.25\\
  \bottomrule
\end{tabular}
\end{table}

\begin{figure}[h]
  \centering
  \includegraphics[width=0.8\linewidth]{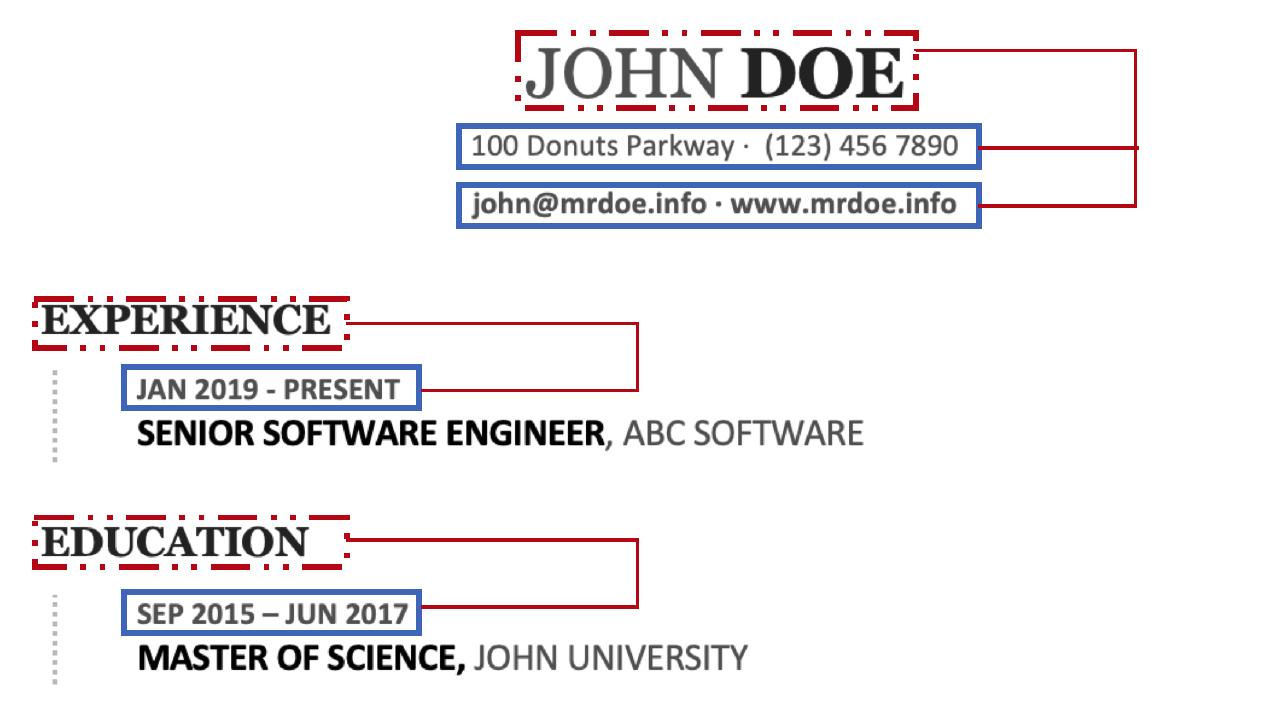}
  \caption{A resume example}
  \Description{An resume example.}
\label{fig:resumesample}\end{figure}
\begin{table}[h]
  \caption{Model accuracy on Resume Dataset}
  \label{tab:resumedataf1}
  \begin{tabular}{llc}
    \toprule
    Model&Fine-tuning task&F1\\
    \midrule
    BiLSTM-CRF & - & 62.71\\
    BERT & - & 67.53\\
    RoBERTa & - & 67.40\\
    \midrule
    RoBERTa+GCN & - & 71.31\\
    \midrule
    RoBERTa+GCN & +SPRC & 71.43\\
    RoBERTa+GCN & +MLM & 71.68\\
    RoBERTa+GCN & +MLM+SPRC & 72.13\\
  \bottomrule
\end{tabular}
\end{table}

\begin{table*}[h]
 \newcommand{\tabincell}[2]{\begin{tabular}{@{}#1@{}}#2\end{tabular}}
  \caption{Statistics and accuracy of the tags in Resume Dataset}
  \label{tab:resumetagdata}
  \begin{tabular}{cccccccccccc}
    \toprule
     & Degree & Position & School & Name & \tabincell{c}{Company\\Duration} & Email & \tabincell{c}{School\\Duration} & Company& Phone & \tabincell{c}{Section\\Title} & Address\\
    \midrule
    RoBERTa F1  & 73.15 & 68.33 & 68.69 & 70.92 & 70.38 & 91.29 & 52.22& 56.53 & 83.57 & 68.52 & 72.07\\
    RoBERTa+GCN F1  & 72.90 & 67.99 & 71.21 & 76.54& 76.79 & 92.37 & 75.43& 59.65& 81.69 & 72.24 & 72.75\\
    \bottomrule
\end{tabular}
\end{table*}
\subsection{Resume IE}

We test on another real-world dataset containing hundreds of international resumes which are all in different layouts. This dataset is different from the invoice dataset in both text content and layout and can be considered as a different task. Extensive experiments are conducted on this dataset to evaluate if the proposed model can generalize across different types of documents.

\subsubsection{Dataset}
The resume dataset consists of 472 labeled documents and 2,130 unlabeled documents. A sample of the dataset is illustrated in Figure~\ref{fig:resumesample}. Compared with the invoice dataset, layout of resumes is more diverse. Text boxes in a page are sometimes arranged in a single column, sometimes double columns, and sometimes even irregular shapes.

The 472 labeled documents contain 1,281 pages and 52,882 text boxes in total. We annotate and extract 11 entity types such as Name, School etc. As the layout of the resumes are diverse enough themselves, we do not perform the few-shot experiment on this dataset.

\subsubsection{Implementation Details}
Different from the invoice dataset that the graph only have one kind of edges, there are two kinds of edges in graph module for resume extraction. One represents the nearest neighbor relationship and the other represents the relationship of each text box and its section title. By analyzing the data set, we notice that there is an obvious layout pattern that is ubiquitous in resume documents: the \textit{section title}, like WORK EXPERIENCE in Figure~\ref{fig:resumesample}. Therefore, we try to encode the relationship of one text box and its section title with a graph network to better use visual information into the model. For example, in Figure~\ref{fig:resumesample}, the green lines represent the adjacent connections and the orange lines represent the section title connections.

In this work, we use a simple rule to help the text boxes find their section title. We assume that the section title of a text box is its nearest neighbor whose bounding box vertical position is above the current box and has a larger font size.

\begin{figure}[h]
  \centering
  \includegraphics[width=0.65\linewidth]{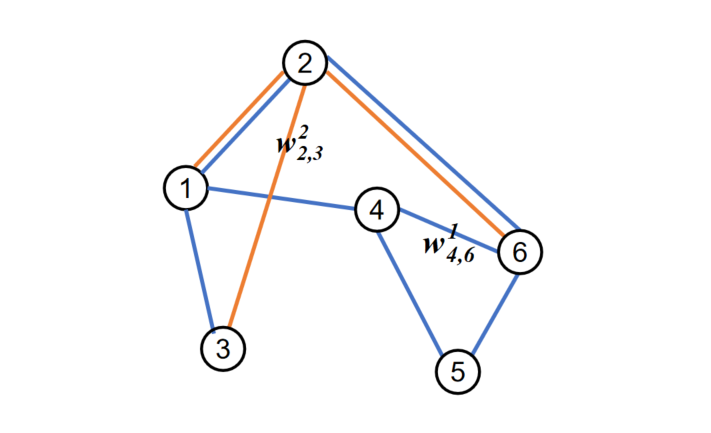}
  \caption{Graph edges setting}
  \Description{A sample of graph edge.}
\label{fig:edge}\end{figure}

GCN in our settings encodes multiple edges follow the Eq (5) that no additional nodes are added. Figure~\ref{fig:edge} shows an example of a graph when there are multiple edge types in the graph.

As for the fine-tuning objective SPRC, the layouts of resume documents are different from those of invoice documents, with very unbalanced number of four relationship types for the classification task. The number of vertically adjacent pairs is about ten times the number of horizontally adjacent pairs. Therefore, we randomly sample one-tenth of the pairs with up-down or down-up labels. The training dataset for SPRC objective is consist of these sampled pairs and the other two types of pairs.
\subsubsection{Dataset}
For the experimental setup, we keep the same experimental settings as the invoice extraction experiment, except that we set the hidden size of GCN to 512 and the maximum node number to 150.
\subsubsection{Results}
The evaluation performance of baselines and our methods on resume dataset are shown in Tabel~\ref{tab:resumedataf1}, which are highly consistent with previous experimental results on invoice extraction. For the graph module experiment, RoBERTa with GCN outperforms the baseline by 3.9\%. In more detail, F1-scores for all the entities are listed in Tabel~\ref{tab:resumetagdata}. It is obvious that our model significantly improves the performance on extracting position sensitive entities such as School Duration (which is easy to be confused with Company Duration, if no layout information is considered) and Section Title.

Fine-tuning objectives are also applied to this problem. As illustrated in Table~\ref{tab:resumedataf1}, both training objectives improve the model performance, and fine-tuning with MLM and then SPRC obtains the highest F1 score at 72.13 on this dataset.

An ablation study on the graph module is conducted to explore the contributions brought by the section title connections, the fonts features and the skip connections. In Table~\ref{tab:ablation}, we observe a 0.8 points drop when the section title edge type is ablated. We also observe that combining the fonts features with each node leads to a 0.5 points improvement over the graph module. Furthermore, dropping the skip connections between graph layers result in a 0.4 points decrease. These results demonstrate that all types of layout information, from position to font, contribute positively to extraction performance.

\begin{table}
  \newcommand{\tabincell}[2]{\begin{tabular}{@{}#1@{}}#2\end{tabular}}
  \caption{Ablation study on graph module}
  \label{tab:ablation}
  \begin{tabular}{lc}
    \toprule
    Model & F1\\
    \midrule
    RoBERTa+GCN (full) & 71.31\\
    RoBERTa+GCN w/o section title edges & 70.55\\
    RoBERTa+GCN w/o fonts feats & 70.81\\
    RoBERTa+GCN w/o section title edges \& fonts feats& 69.99\\
    RoBERTa+GCN w/o skip connections & 70.87 \\
  \bottomrule
\end{tabular}
\end{table}

\section{Conclusions and Future Work}
We introduced a novel approach for structural-aware IE from visually rich documents with fine-tuning objectives which are proved to be both effective and robust. We use GCN to encode various rich layout information and transformer-based pre-trained language models to encode text information. We design two fine-tuning objectives to fully utilize unlabeled data and reduce annotation cost. Experimental results on two datasets and on the few-shot setting suggest that incorporating rich layout information and expressive text representation significantly improves extraction performance and reduces annotation cost for information extraction from visually rich documents.

In future, we plan to explore better integration of text and layout feature for information extraction on business documents. Possible directions include incorporating more spatial and formatting features and better modeling of the relationships between text boxes.

\bibliographystyle{ACM-Reference-Format}
\bibliography{vrd_ie}

\end{document}